# Technical Report: Quality Assessment Tool for Machine Learning with Clinical CT

Riqiang Gao[a]*, Mirza S. Khan [b,c,d], Yucheng Tang[a], Kaiwen Xu[a], Steve Deppen [b], Yuankai Huo[a], Kim L. Sandler [b], Pierre P. Massion[b,d], Bennett A. Landman[a, b]

[a]Electrical Engineering and Computer Science, Vanderbilt University, Nashville, TN, USA 37235
[b]Vanderbilt University Medical Center, Nashville, TN, USA 37235
[c]Department of Biomedical Informatics, Vanderbilt University, Nashville, TN, 37212
[d]U.S. Department of Veterans Affairs, Nashville, TN, 37212
(*corresponding author: riqiang.gao@vanderbilt.edu)

**Abstract**

Image Quality Assessment (IQA) is important for scientific inquiry, especially in medical imaging and machine learning. Potential data quality issues can be exacerbated when human-based workflows use limited views of the data that may obscure digital artifacts. In practice, multiple factors such as network issues, accelerated acquisitions, motion artifacts, and imaging protocol design can impede the interpretation of image collections. The medical image processing community has developed a wide variety of tools for the inspection and validation of imaging data. Yet, IQA of computed tomography (CT) remains an under-recognized challenge, and no user-friendly tool is commonly available to address these potential issues. Here, we create and illustrate a pipeline specifically designed to identify and resolve issues encountered with large-scale data mining of clinically acquired CT data. Using the widely studied National Lung Screening Trial (NLST), we have identified approximately 4% of image volumes with quality concerns out of 17,392 scans. To assess robustness, we applied the proposed pipeline to our internal datasets where we find our tool is generalizable to clinically acquired medical images. In conclusion, the tool has been useful and time-saving for research study of clinical data, and the code and tutorials are publicly available at https://github.com/MASILab/QA_tool.

## 1. Introduction

Medical imaging plays an important role in medicine. It is widely used in clinical diagnosis and for screening programs such as National Lung Screening Trial (NLST) [1] and Vanderbilt Lung Screening Program (VLSP, https://www.vumc.org/radiology/lung). Moreover, medical imaging analyses using artificial intelligence have advanced dramatically in recent years, especially with the development of deep learning [2]–[4].

The use of medical imaging for research and discovery has become more prominent and has led to meaningful contributions to imaging and medicine [3][5][6]–[8]. Often, analyses relying on medical imaging utilize the raw Digital Imaging and Communications in Medicine (DICOM) image files or other common image file formats, such as Neuroimaging Informatics Technology Initiative (NIfTI). In computed tomography (CT), scans are often represented as collections of two-dimensional images (e.g., DICOM) that must be re-composed / re-formatted to determine three-dimensional structure or higher (e.g., NIfTI).

Factors such as network issues, accelerated acquisitions, motion artifacts, and imaging protocol design can impede interpretation of collections of two-dimensional images, especially under the context collaboration of multiple teams [9]. For example, the NLST dataset [1] integrates CT images from over 30 sites across over 20 manufacturer and model combinations, which may trigger data issues when combined. Even within a single institution, such as the Vanderbilt University Institute for Imaging Science Center for Computational Imaging (VUIIS CCI) [10] database, issues such as incomplete transfer may happen when the workload is heavy. Moreover, artifacts such as ring artifacts, motion artifacts and metal artifacts are commonly encountered in clinical CT [11]. Potential data quality issues can be exacerbated when the data collection pipeline includes human-based workflows while use limited views of the data at each step. In medical image de-identification, minor encoding errors or manipulation of DICOM file data could render these files useless or inaccessible. Data quality issues may obscure digital artifacts that can drive inference toward erroneous decisions. An intuitive example shown in Figure 1, when analyzing the lung cancer through CT image, the artifacts on pulmonary nodule may cause the misleading decision for doctors and encode biased information for machine learning algorithm.

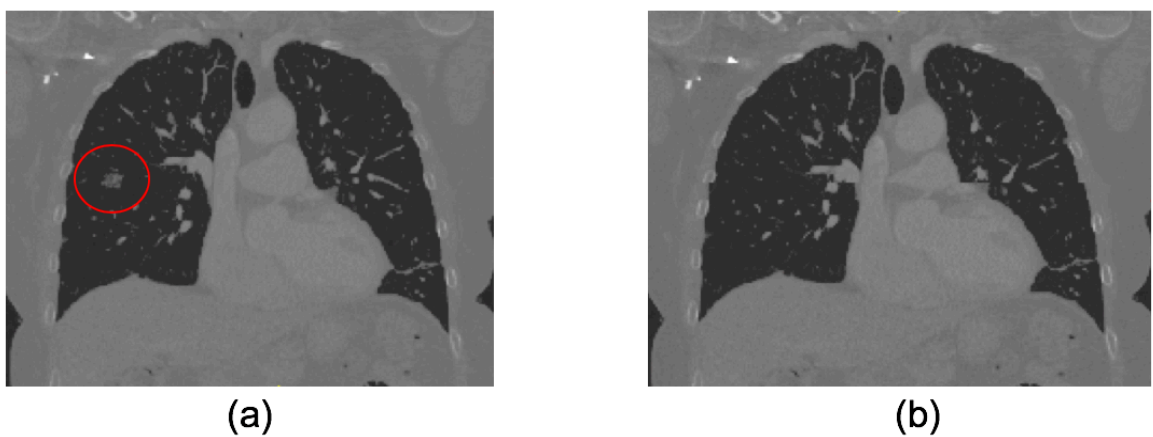

**Fig. 1.** Potentially misleading interpretation of an image with slices missing. (a) The complete image with a pulmonary nodule highlighted within the red circle. (b) The same image with a few slices missing. The image from (b) can be misinterpreted as lacking a pulmonary nodule to human reviewers and AI algorithms alike.

Image quality assessment (IQA) can be divided into two categories [12]: (1) subjective evaluation (assessed by a human reviewer) and (2) objective assessment (computed by algorithms) including machine learning based methods [13][14]. The subjective assessment is usually regarded as the gold standard for IQA. Objective assessment of low contrast detectability in CT images was proposed in [15]. Medical imaging quality noises from acquisition of full reference based IQA and no reference IQA are discussed in [16]. Deep learning has also been applied to CT image quality assessment [17]. The subjective assessment is reliable, requires a lot of time and human resources. The objective assessment may be more efficient, but accuracy cannot be assured. Moreover, the IQA is especially unstable for cases not seen in the training as a common drawback of machine learning algorithms. In practice, multiple reasons can result in the image quality concerns. Some of cases can be fixed by dealing with the image itself and some cannot; this further increases the efforts of subjective assessment and decreases the robustness of objective assessment. A comprehensive IQA pipeline that integrates the advantages of subjective and objective assessment for effectiveness and robustness is needed.

Quality control is essential in medical image data handling [18], either during acquisition (e.g., [11][19]) or post-processing (e.g., [20]). Branco et al. [19] used an updated credence cartridge radiomics phantom to detect subtle artifacts. The DTIPrep tool provides a pipeline with several quality control steps with a detailed protocoling and reporting facility for diffusion weighted images. [20]. While the quality of images still need to assessed before clinical or research use [21]. As one of the largest publicly available chest CT datasets, NLST has been widely used in numerous medical imaging research studies. The reporting on IQA with the NLST has been varied: Ardila et al. [3] excluded the volumes that failed to download or had unparseable DICOMS, and removed the volumes with (1) fewer than 50 slices, (2) slice spacing $\geq$ 5.0 mm, and (3) inconsistent pixel spacing. Schreuder et al. [22] skipped the CT images of corrupted images or incomplete images. Gierada et al. [23] conducted a subject review of NLST using 3.1% subjects in NLST, which is time-consuming. However, the studies in [3][22] do not detail how they specifically performed IQA. Researches based on NLST are from multiple tasks including lung cancer risk prediction [24]–[27], lung cancer incidence or mortality analysis [28][29], vertebra segmentation [30], coronary artery calcium scoring [31], but no description of QA can be found in these references.

The VUIIS CCI [10] database was developed to provide a new framework for algorithm development allowing large scale batch processing of images and scalable project management, which is built on XNAT open source imaging informatics platform [32]. Additional de-identified research efforts involving chest CT, such as the Molecular and Cellular Characterization of Screen-Detected Lesions (MCL, https://mcl.nci.nih.gov/) and Vanderbilt Lung Screening Program (VLSP, https://www.vumc.org/radiology/lung) datasets are stored in VUIIS CCI. Unable to exclude the potential for noise or errors during medical imaging acquisition and/or data transfer, we find that IQA is a vital process for data stored in VUIIS CCI. As more medical imaging datasets become available, researchers must be cognizant of the potential for faulty imaging studies being present. Furthermore, for the purposes of replicability, a standardized QA method is necessary.

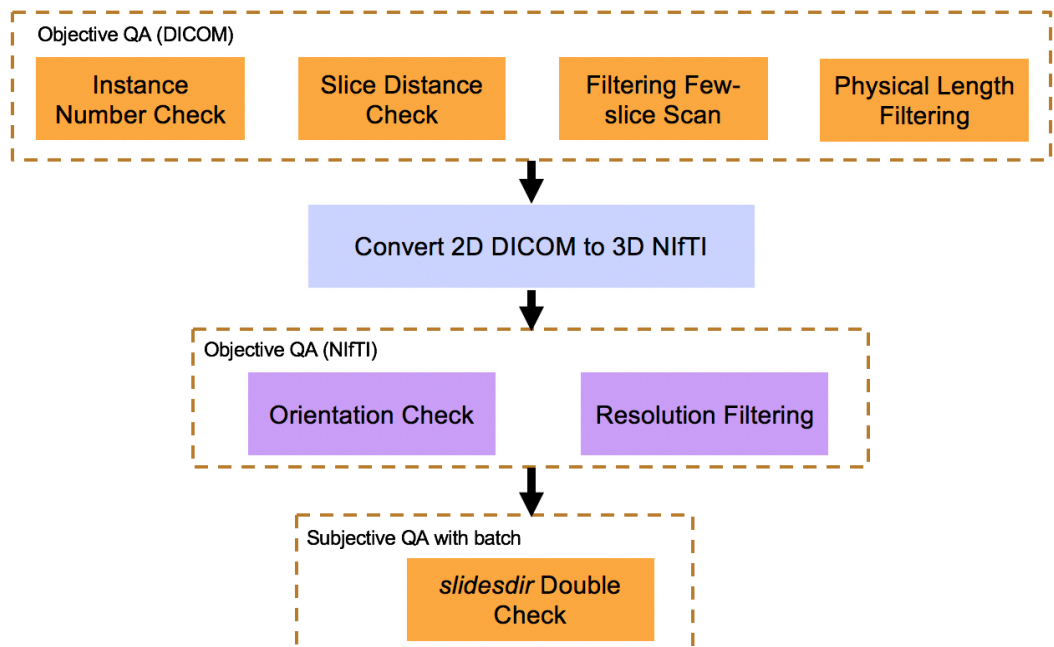

**Fig. 2**. The components in our IQA tool pipeline. Objective assessment on DICOM and NIfTI, subjective assessment with batch are included.

To address IQA for clinical research and our own model development, we developed a series of automated data quality checks for medical imaging data, as shown in Figure 2. Our IQA tool is based on the Python platform, which integrates the advantages of subjective and objective assessments. The IQA tool can identify the issues from slice missing, low axial resolution, out of Region of Interest (ROI), and provides a comprehensive report. First, our tool performs objective QA on DICOM images. Then, the DICOMs are converted to NIfTI and another objective QA is performed. The QA report can be automatically generated for each objective check, which is more efficient than human-only review. Finally, the NIfTI files are checked by subjective QA in bulk using the *slicesdir* tool for a more focused manual review. Our IQA tool is flexible, allowing users to select specific QA steps as desired. Additionally, our tool not only provide steps to identify certain errors, but also provide options to correct the image if it is possible with its own. We demonstrate its usage on NLST and clinical imaging studies (MCL and VLSP). The code and tutorials are publicly available at https://github.com/MASILab/QA_tool.

## 2. Related Public Tools used in our Pipeline

*dcm2niix* [33]: *dcm2niix* is an open-source tool that is designed to convert neuroimaging data from the DICOM to NIfTI format.

*slicesdir* [34]: *slicesdir* is tool that takes in a list of images, and for each one, runs slicer to produce the same 9 default slices and combine them into a single GIF picture. All the GIF pictures can be viewed through one webpage. *slicesdir* belong to the family of FSL [34]. FSL is a comprehensive library of analysis tools for medical imaging.

*fslreorient2std* [34]: this is a tool to reorient an image to match the orientation of the standard template images (MNI152) [35] so that they appear "the same way around" in FSLView [34].

## 3. Detailed Steps

We first describe the several steps of quality assessment (QA) for the de-identified medical images, including the steps converting 2D DICOM files to 3D NIfTI files.

### 3.1 Instance Number checking

*Instance Number (IN)* is a number that identifies this image in DICOM files, which was named Image Number in earlier versions [36]. Generally, the INs of DICOM images in a single scan should be a series of consecutive integers. Our first priority is to check if there are any missing INs of the CT scan. The rationale behind this step is that slices are missing when the number of slices is fewer than the number of the maximum IN minus the minimum IN plus one.

We check the IN using

$$C_1 = max\{in_i\} - min\{in_i\} + 1 - size\{in_i\}$$

$$C_2 = \sum_{i,j,i<j} \mathbf{1}\{in_i == in_j\}$$

where $\{in_i\}$ is the list of INs of the DICOM files, and $size\{in_i\}$ indicates how many DICOMs in the scan. $C_1$ and $C_2$ represent how many slices are missed in the scan and how many slice-pairs with the same IN, respectively. The CT scan pass the Instance Number checking if $C_1 = 0$ and $C_2 = 0$.

### 3.2 Slice Distance checking

*Slice Location* is defined as the relative position of the image plane expressed in mm. This information is relative to an unspecified implementation specific reference point [36]. *Slice Distance (SD)* is defined as subtraction of *Slice Location* of the two consecutive DICOM image files. This step is introduced to check the SD between all consecutive DICOM files in their ordered sequence. Generally speaking, the SD of all consecutive pairs should be the same in a scan, while in practice, the system error might result in very small and almost negligible difference. Herein, we introduce a self-defined threshold to specify our tolerance for slice distance variation.

The SD checking is

$$C_3 = \sum_i \mathbf{1}\{sd_i < \varepsilon\}$$

where $\{sd_i\}$ is the list of SDs of DICOM files, $\varepsilon$ is the self-defined tolerance threshold, which should be related to the axial resolution. $C_3$ represents how many places which slice distance error. The CT scan pass the Slice Distance Check if $C_3 = 0$.

### 3.3 Filtering scans with few slices

Some scans with an unreasonably limited number of slices may pass the Instance Number and Slice Distance checks. For example, if a chest scan only with three DICOM slices, it is unlikely to be usable. This step filters few-slice scans.

Filtering such scans can be done with a user-defined threshold on $size\{in_i\}$.

$$C_4 = \mathbf{1}\{size\{in_i\} < \delta\}$$

where $\delta$ is the self-defined threshold for filtering scans with a limited number of slices. The CT scan pass the filtering checking if $C_4 = 0$.

### 3.4 Physical Length Filtering

Some scans may extend outside of the region of interest (ROI). For example, the target scan is chest CT while is given a whole-body scan may have been provided. The Physical Length Filtering is designed for selecting those CTs with problematic physical body length for further processing or removal.

Two thresholds are needed (i.e., low bound and upper bound of physical body length) for filtering

$$C_5 = \mathbf{1}\{\sigma_1 < PL < \sigma_2\}$$

where PL is the physical length computed from Slice Location of DICOMs, $\sigma_1$ and $\sigma_2$ are the self-defined lower bound and upper bound. The CT scan passes the filtering checking if $C_5 = 1$.

We provide two additional steps to segment the ROI around the lung, as shown in Figure 3.
Step 1. Create the lung mask based on the preprocessing of [37].
Step 2. Segment the lung ROI with lung mask.

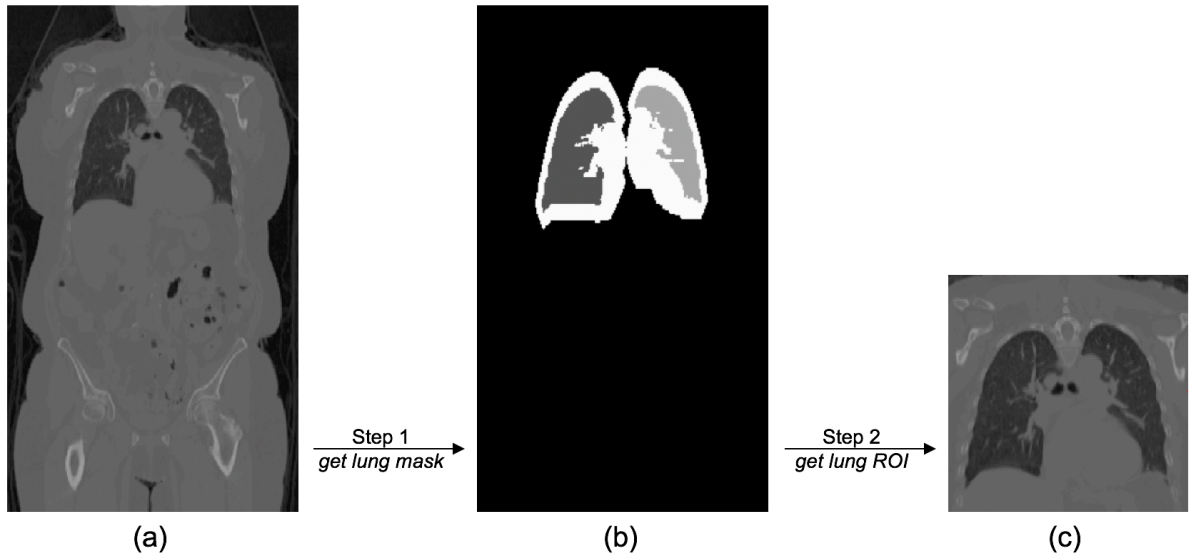

**Fig. 3.** The two steps to segment scan that out of ROI. The Step 1 (getting lung mask) is adapted from [37], which is based on thresholding, dilation. The Step 2 is cropping the image based on the lung mask with user-defined margin extension (e.g., 10%).

### 3.5 NIfTI orientation Check and Resolution filtering

Image orientation is an important issue, but may appear confusing [34]. This step is introduced to check and re-orient the image (if necessary).

After converting DICOM files to the NIfTI file format using an open-source tool *dcm2niix* [33], we check the CT orientation and resolution in the affine matrix *A*, whose size is 4x4. We have

$$C_6 = \mathbf{1}\{-A_{11} == A_{22} == A_{33} > 0\}$$

$$C_7 = \sum_{i=1}^{3} \mathbf{1}\{|A_{ii}| > \Phi_i\}$$

where $\Phi$ is the thresholds for each of the three dimensions. The CT scan is not in standard orientation if $C_6 = 0$, and we use the open source *fslreorient2std* [34] to convert CT to standard orientation, as Figure 4 shows. $C_7 > 0$ suggests that the CT fails to match resolution requirements.

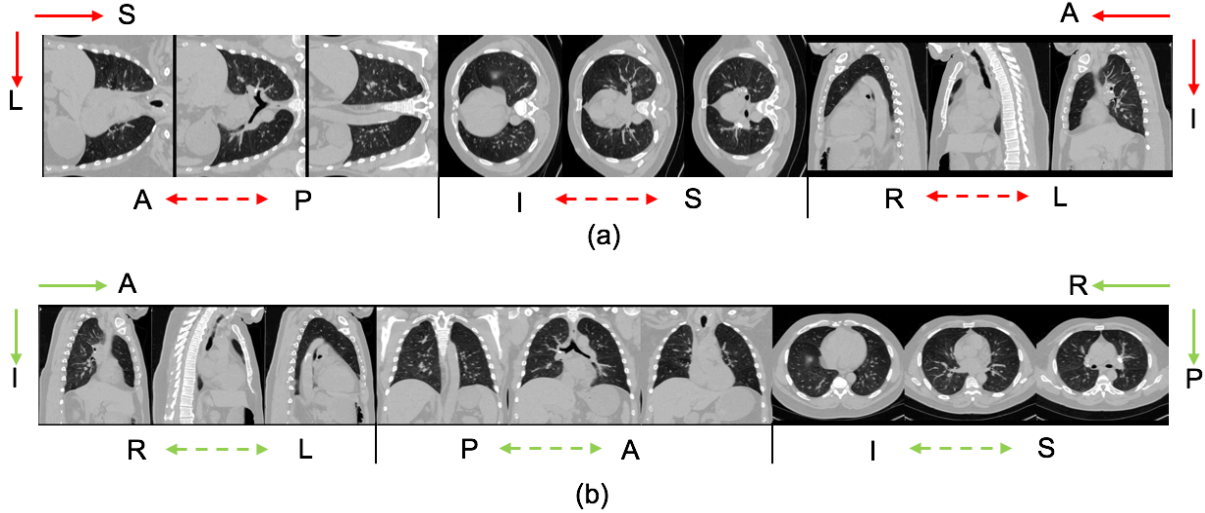

**Fig. 4.** The nonstandard orientation case re-oriented by *fslreorient2std* tool, the scans are displayed by *slicesdir* [34]. (a) nonstandard orientation case, (b) standard orientation case. The dash lines represent the direction of across slices. The full lines represent the CT direction of one single slice. The red and green lines represent the direction of nonstandard and standard cases, respectively. The Anterior (P), Posterior (P), Inferior (I), Superior (S), Right (R), Left (L) annotations follow the guides of [38].

*3.6  Double check with visualized slices*

The subjective assessment is regarded as the gold standard for IQA. In the final step, we apply the subjective assessment for validation using a batched approach, which is much faster than conventional subjective assessment.

We use the *slicesdir* [34] to visualize scans. As described in Section 2, the *slicesdir* tool can create a webpage for manual QA of a large batch of images.

**4.  A case study with the NLST**

*4.1  Data*

National Lung Screening Trial (NLST) [1] is a randomized controlled clinical trial of screening tests for lung cancer using low dose chest computed tomography (CT), which is the largest chest CT dataset publicly available. The goal of the NLST study was to assess whether screening with low-dose CT could reduce mortality from lung cancer [39].

The NLST dataset was downloaded from the official website (https://cdas.cancer.gov/nlst/) using the provided download instructions. We downloaded the JNLP file from the CDAS of Project NLST-7 and then used the Java Web Start Software to download the DICOM files. These DICOM images were stored in our local file system.

*4.2 Results*

QA results from the subsets of NLST dataset are shown in Table 1, with the results included only those CTs for which a diagnosis was ascertained, a total of 17392 scans. Note that (1) some scans may fail in multiple checks, so the total failed cases would be lower than summation of numbers in all checks, (2) some QA-failed cases may only indicate potential warnings and some still can be used in machine learning algorithms or clinical readers. For example, some scans with duplicated DICOMs, but those duplicated DICOMs can be ignored with *dcm2niix* tool to convert to NIFTI, (3) Figure 5 shows the result of the scan with largest number of DICOMs if multiple scans existed in one session. Some sessions may fail in the scan with largest-number-of-DICOMs, while still with usable scan, (4) the number of "Subjective QA with batch" is obtained by review of picked 1,000 randomly selected scans. The Subjective QA with batch is performance by the authors who has no clinical training and experiences. We notice some subjective QA failing cases (e.g., with noise background) can be improved by pre-processing of [37].

Figure 5 shows the axial field of view (FOV) and resolution distributions in NLST. Figure 5.1 shows that the majority of axial FOV are located in the range of 250 mm – 400 mm. The distribution of resolutions is primarily in the range of 1 mm – 3 mm, as expected.

**Table 1**. The IQA results on NLST with confirmed-diagnosis record

|  |  |  |
|---|---|---|
| Objective QA (DICOM) | Instance Number Check | 0.4% |
|  | Slice Distance Check | 3.9% |
|  | Filtering Few-slice Scan | 0.2% |
|  | Physical Length Filtering | 0.25% |
| Objective QA (NIfTI) | Orientation Check | 0.14% |
|  | Resolution Filtering | 0.99% |
| Subjective QA (batch) | Slicesdir Double Check | 0.6% |

**A case study with the MCL/VLSP**

*4.3 Data*

We consider two in-house clinical lung CT datasets in the evaluation: (1) The Molecular and Cellular Characterization of Screen-Detected Lesions (MCL, https://mcl.nci.nih.gov) and (2) Vanderbilt Lung Screening Program (VLSP, https://www.vumc.org/radiology/lung), in total 3029 subjects with 5274 scans. These datasets are stored in the VUIIS CCI database [10]. To illustrate the need for QA, we present analyses prior to our regular QA processes.

*4.4 Results*

Table 2 shows the QA results from the subsets of MCL/VLSP dataset, in total 5274 scans. As with the NLST dataset, (1) some scans may fail in multiple checks and (2) some QA-failed cases still can be used in AI algorithm or clinical usage (e.g., some header issues can be corrected without changing image intensity volume) which can be validated by subjective assessment. The majority of failing cases in "Instance Number Check", "Slice Distance Check" and "Filtering Few-slice scan" have been fixed by re-transferring the data. We show the original number here to illustrate that a number of errors can occur in practice.

Figure 6 shows the axial FOV and resolution distributions of in-house datasets. Compared with our evaluation on NLST, we find that the in-house datasets are with larger variations. The potential reasons are (1) the NLST dataset is collected with more strict inclusion/exclusion criteria, (2) the in-houses datasets combine multi-site data and cross large range of collecting years (> 5 years).

**Table 2**. The IQA results on our in-house datasets

|  |  |  |
|---|---|---|
| Objective QA (DICOM) | Instance Number Check | 8.4% |
|  | Slice Distance Check | 10.3% |
|  | Filtering Few-slice Scan | 4.5% |
|  | Physical Length Filtering | 4.8% |
| Objective QA (NIfTI) | Orientation Check | 0.3% |
|  | Resolution Filtering | 8.5% |
| Subjective QA (batch) | Slicesdir Double Check | 0.3% |

*Note that most of failing cases of "Instance Number Check", "Slice Distance Check", "Filtering Few-slice Scan" are caused by the data transferring, which can be fixed by re-transferring the data through XNAT again.

*4.5 Pilot Study on impact with machine learning*

As we mentioned, we find data slices can be missed during data transfer. To demonstrate the effect on learning from erroneous scans, we selected 5 chest CT scans of patients with known lung cancer from MCL. We compare the predicted cancer probability between the complete scans (which completed by re-transfer) and scans where some slices are missing. The predicted cancer probabilities are computed by the state-of-the-art method: Liao et al. [37] with their public pre-trained model.

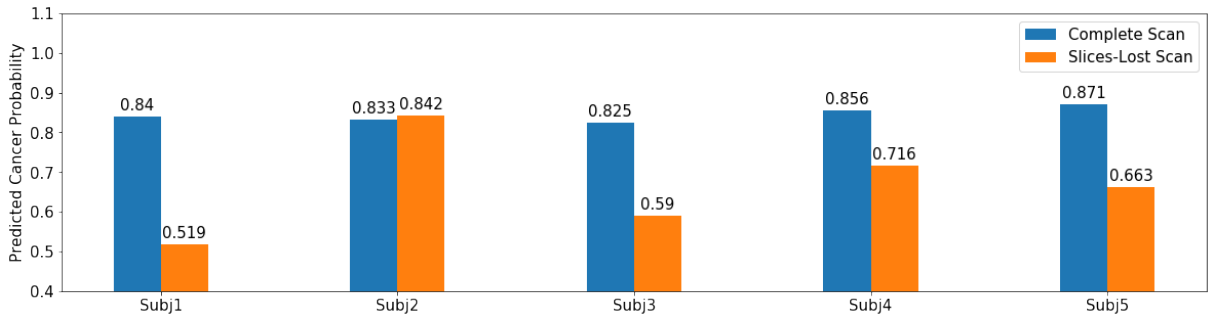

**Fig. 7.** The comparison of predicted cancer probability of complete CT and slices-lost CT. The scans are all from cancer patients. The slices-lost is mainly caused by data exchange across multiple platforms, and re-transfer the data can make the CT complete.

As Figure 7 shows, the predicted cancer probability may change dramatically when the CT image loses slices. This indicates that the images that fail QA may adversely affect the machine learning process.

## 5. Detected Problematic Examples

*5.1 Case of instance number failing*

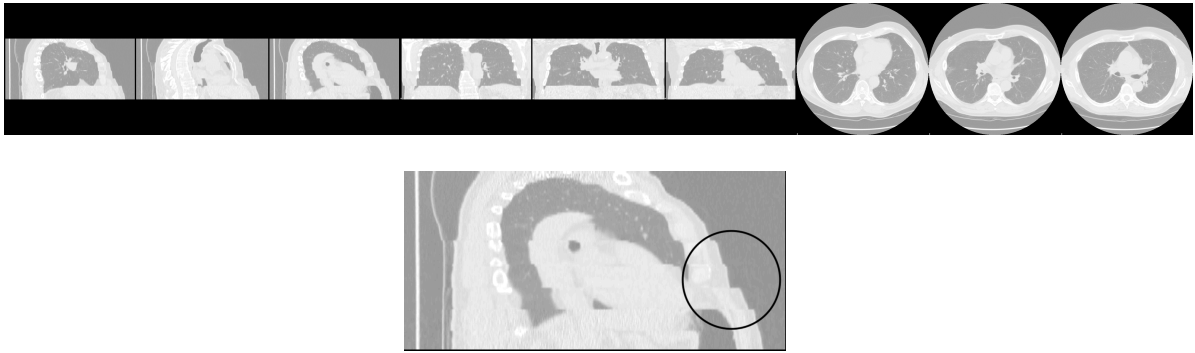

**Fig. 8.** Failing case of instance number check. Some slices are missed this chest CT, which can be indicated by the black circle of the bottom panel.

Figure 8 illustrates a case where slices are missed in the CT scans. As an intuitive example in Figure 1, this kind of error can lead to losing important phenotype for diagnosis.

## 5.2  Case of Slice Distance failing

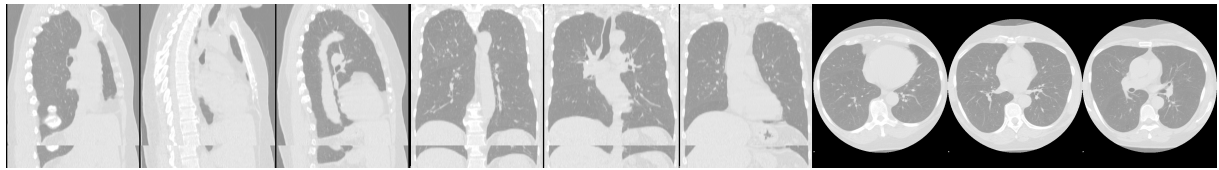

**Fig. 9.** Failing case of slice distance check. Duplicated chunk of slices is illustrated in the bottom of CT.

Figure 9 shows duplicated chunk of CT in the scan, which can be detected by the *Slice Distance Check*. This error can make the preprocessing of image fail before processing with machine learning algorithm.

## 5.3  Case of Filtering Too Few Slices

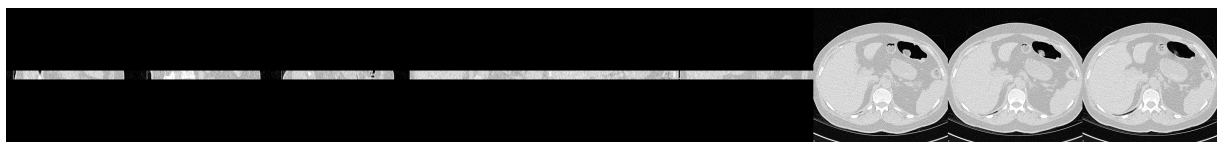

**Fig. 10.** Case of the Filtering few slices. The number of slices is less than 20.

A too-few-slices-scan case is shown in Figure 10. This error can result from data acquisition or data exchange. This kind of cases cannot provide complete information (even misleading) for machine learning or clinical usage.

## 5.4  Case of Physical Length Filtering

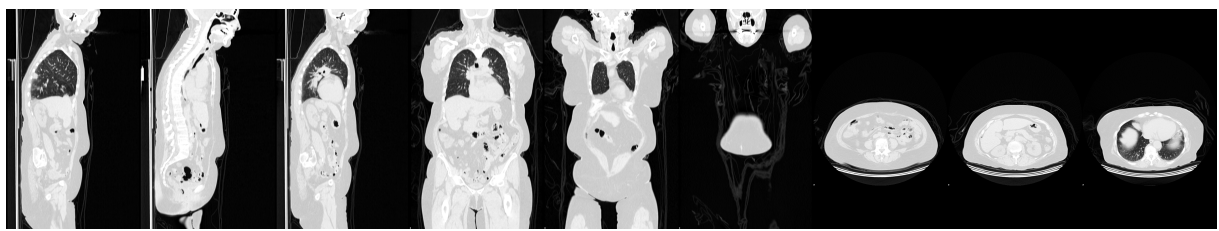

**Fig. 11.** Case of the Physical Length Filtering. The region of CT extends the region of chest.

Some scans have larger typical ROI, as shown in Figure 11. Generally, this kind of scans still can be used. One option is to segment the lung region based on our tool, as Figure 3 shows.

## 5.5  Case of Orientation Check

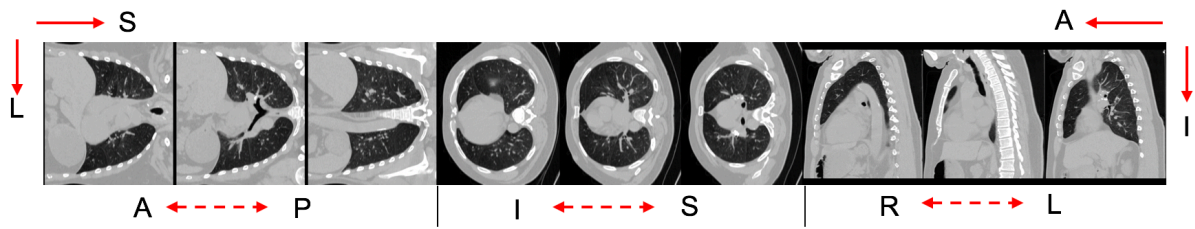

**Fig. 12**. Case of orientation check. The CT is not in a standard orientation.

A few cases in both NLST and our in-house dataset has a nonstandard header orientations (Figure 12). The orientation can cause downstream process to create the wrong data matrices when loading to machine learning algorithm, which should be harmful for learning. The orientation error can be fixed use the *fslreorient2std* [34] tool, which has been used in our pipeline.

## 5.6 Case found in slicesdir Double Check

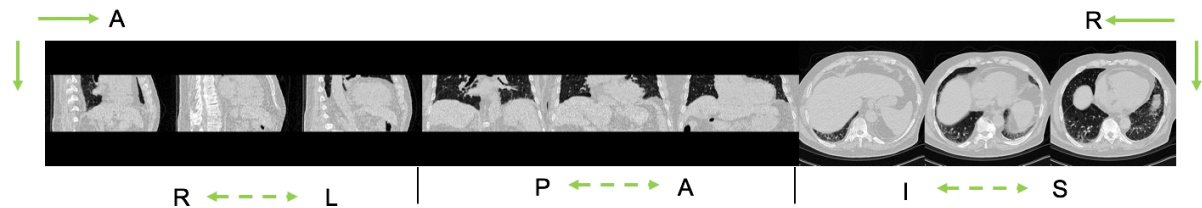

**Fig. 13**. Failing case found by *slicesdir*. The chest CT does not contain a complete lung.

Even though most of the problematic cases can be detected by above objective QA steps, a few cases still are not with high quality for next step analysis. As shown in Figure 13, the ROI has not been covered by the scan. In our pipeline, we include the subjective QA with large batch based on the tool *slicesdir* [34]. Subjective QA with batch is much faster than the one-of-a-time subject QA.

## 6. Discussion

Even though the subjective review is considered the gold standard for QA, subjective review of each medical image is difficult with large-scale clinical datasets. Also, missing slices may falsely appear correct on visual inspection, as highlighted in Figure 1, which can deceive reviewers in a busy practice. Given the unknownness, heterogeneity, and variability in sources of error, objective and automated assessment is not without limitation. Our tool combines the advantages of subjective and objective assessments to provide an efficient QA method. In Figure 8 and 13, we show that low-quality scans can be both challenging for clinical usage and AI algorithm. Our tool can effectively reduce those side-effects.

In this paper, we introduce a QA tool for CT images with multiple steps, including objective and subjective assessment. There are multiple steps for checking quality in our tool and each step can give a different aspect of understanding of data. This process can help to identify the reason for failing QA and determine to fix the image or discard the image if cannot be fixed. The Instance Number Check aims to find how many slices are missed by checking the DICOM headers. Slice Distance Check can filter out the scans with duplicated chunks. The acquisition error resulting in few slices for one scan can be detected by Filtering Few-Slice Scan. The extending ROI case can be detected by the Physical Length Filtering, and further steps (e.g., segment the lung) can be processed accordingly. The inconsistent orientations problem may be harmful to machine learning since the data matrices are not standardly oriented, which can be detected and corrected by the Orientation Check step. As the CT scans can be acquired by different settings, the Resolution Filtering can select the user-defined resolution range. Finally, a quick batch-based subjective QA process is achieved by slicesdir Double Check.

NLST is a well-known large-scale dataset with chest CTs. The previous QA tools on NLST either applying few objective criteria to filter obvious errors [3][22], or fully subjective QA requires substantial human efforts [23]. In addition, compared with NLST which is quite standardized and large human efforts have spent on the collecting and assessment, in-house clinical datasets have more issues on CT image quality, especially when managing data from multiple sites. Our tool seeks a balance between the efficiency and complexity by including several objective checks and subjective check in batches. Also, our tool allows users to adjust the efficiency / complexity balance by flexible choosing the certain steps to apply.

Our study has some limitations. First, our tool mainly focuses on the CT images already obtained, and we do not analyze the parameters of scanner. Second, our tool is not based on automatic image learning context, thus, the artifacts such as ring artifacts, motion artifacts and metal artifacts only can be found with subjective QA step. Third, the intent of our tool is for research studies, the tool cannot intervene in the typical clinical workflows and DICOM standards.

We maintain a publicly available GitHub repository for updating and tutorials at https://github.com/MASILab/QA_tool.

## 7. Acknowledgements

This research was supported by NSF CAREER 1452485 and R01 EB017230. This study was supported in part by a UO1 CA196405 to Massion. This study was in part using the resources of the Advanced Computing Center for Research and Education (ACCRE) at Vanderbilt University, Nashville, TN. This project was supported in part by the National Center for Research Resources, Grant UL1 RR024975-01, and is now at the National Center for Advancing Translational Sciences, Grant 2 UL1 TR000445-06. We gratefully acknowledge the support of NVIDIA Corporation with the donation of the Titan X Pascal GPU used for this research. The de-identified imaging dataset(s) used for the analysis described were obtained from ImageVU, a research resource supported by the VICTR CTSA award (ULTR000445 from NCATS/NIH), Vanderbilt University Medical Center institutional funding and Patient-Centered Outcomes Research Institute (PCORI; contract CDRN-1306-04869). This study was funded in part by the Martineau Innovation Fund Grant through the Vanderbilt-Ingram Cancer Center Thoracic Working Group and NCI Early Detection Research Network 2U01CA152662-06. Mirza S. Khan is supported by the VA OAA Advanced Fellowship Program in Medical Informatics.